\documentclass[journal,twoside,web]{ieeecolor}
\usepackage{jsen}
\usepackage{cite}
\usepackage{amsmath,amssymb,amsfonts}
\usepackage{algorithm}
\usepackage{algorithmic}
\usepackage{graphicx}
\usepackage{textcomp}
\usepackage{wrapfig}
\usepackage{booktabs}
\usepackage{tabularx}
\usepackage{float}
\usepackage{svg}
\usepackage{caption}
\usepackage{graphicx}
\usepackage{subcaption}
\usepackage{float}
\usepackage{subcaption}  

\def\BibTeX{{\rm B\kern-.05em{\sc i\kern-.025em b}\kern-.08em
    T\kern-.1667em\lower.7ex\hbox{E}\kern-.125emX}}
\definecolor{abstractbg}{rgb}{0.89804,0.94510,0.83137}
\begin{document}

\title{DLBAcalib: Robust Extrinsic Calibration for Non-Overlapping LiDARs Based on Dual LBA }

\author{
    Han Ye, Yuqiang Jin,\IEEEmembership{Graduate Student Member, IEEE}, Jinyuan Liu,Tao Li,Wen-An Zhang and Minglei Fu, \IEEEmembership{Member, IEEE}
    \thanks{Manuscript received June 23, 2025; revised XXXX XX, 2025; accepted XXXX XX, 2025. Date of publication XXXX XX, 2025; date of current version XXXX XX, 2025. This work was supported in part by the Key R\&D Program of Zhejiang under Grant 2025C01075, in part by the National Natural Science Foundation of China under Grant U24A20249, in part by the Open Fund of the Technology Innovation Center for 3D Real Scene Construction and Urban Refined Governance, Ministry of Natural Resources (Grant No. 2024PF-3), in part by the Key R\&D Program of Ningbo under Grant 2024Z300, 2023Z220, and in part by the Public Welfare Science and Technology Plan Project of Ningbo under Grant 2024S061.}
    \thanks{H. Ye, Y. Jin, J. Liu, Wen-An Zhang and M. Fu are with the Department of Information Engineering, Zhejiang University of Technology, Hangzhou 310023, China (e-mail: \{hye, yqjin, jy.liu,tao\_li,Wen-An Zhang,fuml\}@zjut.edu.cn).}
}

\IEEEtitleabstractindextext{%
\fcolorbox{abstractbg}{abstractbg}{%
\begin{minipage}{\textwidth}%
\begin{wrapfigure}[12]{r}{3in}
\vspace{-5mm}  
\includegraphics[width=3in]{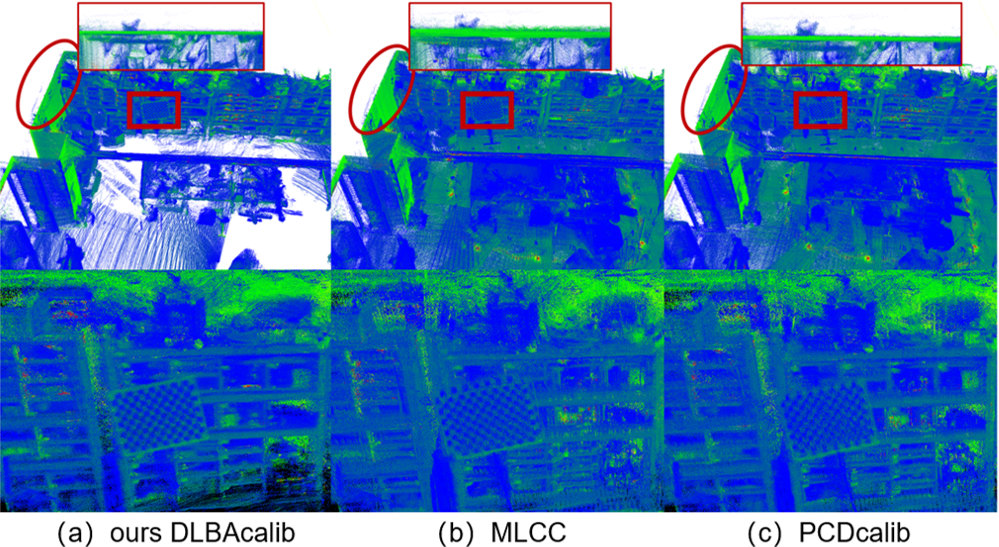}
\vspace{-2mm}  
\end{wrapfigure}

\begin{abstract}

Accurate extrinsic calibration of multiple LiDARs is crucial for improving the foundational performance of three-dimensional (3D) map reconstruction systems. This paper presents a novel targetless extrinsic calibration framework for multi-LiDAR systems that does not rely on overlapping fields of view or precise initial parameter estimates. Unlike conventional calibration methods that require manual annotations or specific reference patterns, our approach introduces a unified optimization framework by integrating LiDAR bundle adjustment (LBA) optimization with robust iterative refinement. The proposed method constructs an accurate reference point cloud map via continuous scanning from the target LiDAR and sliding-window LiDAR bundle adjustment, while formulating extrinsic calibration as a joint LBA optimization problem. This method effectively mitigates cumulative mapping errors and achieves outlier-resistant parameter estimation through an adaptive weighting mechanism. Extensive evaluations in both the CARLA simulation environment and real-world scenarios demonstrate that our method outperforms state-of-the-art calibration techniques in both accuracy and robustness. Experimental results show that for non-overlapping sensor configurations, our framework achieves an average translational error of 5 mm and a rotational error of 0.2°, with an initial error tolerance of up to 0.4 m/30°. Moreover, the calibration process operates without specialized infrastructure or manual parameter tuning. The code is open source and available on GitHub (\underline{https://github.com/Silentbarber/DLBAcalib}).
\end{abstract}

\end{minipage}}}

\maketitle

\section{Introduction}
\label{sec:introduction}
\IEEEPARstart{A}{mong} various perception devices, LiDAR has become one of the most commonly used sensor technologies due to its high-precision 3D measurement capabilities. This technology plays an irreplaceable role in applications such as localization and 3D modeling. In autonomous driving and 3D modeling applications, the inherent limitations of single-LiDAR sensing make the integration of multiple LiDARs essential. However, extrinsic calibration between LiDARs in multi-sensor configurations remains a pivotal challenge impacting overall system performance \cite{b1}.

Calibration between LiDARs consists of internal and external calibration. Internal calibration determines the 3D positions of laser points relative to the LiDAR's coordinate system, which is usually provided by the manufacturer. External calibration, the primary focus of LiDAR calibration, involves determining the relative pose between LiDARs and transforming them into a unified coordinate system \cite{b2}.

Based on the current state of research, external calibration methods can generally be categorized into target-based and motion-based approaches. Target-based methods require explicit artificial markers, such as calibration boards, geometric objects, or reflective landmarks \cite{b3}. Motion-based methods align the motion trajectories of the two LiDARs and typically rely on additional sensors, such as GNSS or IMU, to improve trajectory accuracy. In addition to these approaches, there are various target-free methods that reduce dependence on extra sensors and manual references but still require specific application scenarios \cite{b4}.

Despite significant research progress, two persistent challenges hinder practical implementation: 1) Mandatory overlapping fields of view (FoV) constrain deployment flexibility for narrow-FoV mechanical LiDARs and solid-state variants, and 2) Dependency on high-precision initial parameters imposes impractical constraints for real-world applications. Conventional methods relying on FoV overlap, precise initialization, or dedicated targets prove inadequate for large-scale deployments, often leading to error accumulation and reduced robustness.

To address these issues, this paper proposes a calibration method that requires no artificial reference markers (targetless), no overlapping FoV between LiDARs, and no precise initial extrinsic parameters. Our method operates efficiently with only coarse initial estimates (0.4 m/30° tolerance) while avoiding manual parameter tuning. Our contributions are as follows:
\begin{itemize}
    \item We introduce a novel targetless extrinsic calibration method that utilizes only the cumulative point cloud map generated by one of the LiDARs being calibrated. By iterative frame-to-map alignment, our approach removes the need for overlapping fields of view or accurate initial parameter estimates, greatly simplifying the calibration process.
    \item We formulate the extrinsic calibration problem as a LiDAR bundle adjustment (LBA) task and introduce an iterative optimization strategy. This approach jointly refines intra-source and inter-source point cloud frames to minimize errors in the cumulative map. 
    \item We evaluate our method in simulated and real-world settings using datasets with diverse FoV overlaps and point density. Compared to state-of-the-art techniques, our method delivers high accuracy even without overlapping FoVs or precise initial extrinsics, tolerating initial errors up to 0.4 m in translation and 30° in rotation. For a dual-LiDAR setup, we achieve an average translation error of 5 mm and a rotation error of 0.2°. Additionally, we have open-sourced our ROS-based implementation on GitHub to benefit the robotics community.
\end{itemize}

The rest of this paper is structured as follows: Section \ref{sec:related} reviews related work, Section \ref{sec:method} describes the proposed method, Section \ref{sec:experiments} presents the experiments and results used for validation, and Section \ref{sec:conclusion} concludes the paper.

\section{Related Work}
\label{sec:related}
Extrinsic calibration of multiple LiDAR sensors has garnered significant attention in recent years, with various approaches proposed to tackle its inherent challenges. These methods are typically classified into three categories: target-based, feature-based, and motion-based methods. In this section, we review state-of-the-art techniques within each category, discuss their limitations, and position our work relative to these efforts.

\subsection{Target-Based Calibration}
Target-based methods employ artificial markers, such as checkerboards or reflective markers, to establish correspondences between LiDAR scans. The well-defined geometry of these targets enables high calibration accuracy. For example, Liu \emph{et al.} \cite{b5} developed a method using a planar board with circular holes, while Kim \emph{et al.} \cite{b6} utilized a 3D target with known dimensions. Xue \emph{et al.} \cite{b7} proposed an automatic calibration approach using two non-parallel poles adorned with retro-reflective tape, relying on LiDAR intensity data for pole recognition. Despite their precision, these methods demand meticulous target placement and assume simultaneous visibility across all sensors, rendering them impractical for large-scale or dynamic environments.

\subsection{Feature-Based Calibration}
Feature-based methods leverage natural environmental features—such as planes, edges, or corners—to eliminate the need for artificial targets. Iterative Closest Point (ICP)-based techniques \cite{b8} are widely adopted, aligning point clouds by iteratively minimizing point-to-point distances. However, ICP requires a reliable initial estimate and struggles with limited sensor overlap. To address this, Nie \emph{et al.} \cite{b9} introduced a feature-matching method using edge and plane features, while Tahiraj \emph{et al.} \cite{b10} proposed a semantic alignment approach based on detected objects. Kim and Kim \emph{et al.} \cite{b11} developed a plane-matching technique for structured environments using RANSAC and nonlinear optimization. Similarly, Wei \emph{et al.} \cite{b12} presented a two-stage method involving rough calibration with ground plane features and ICP refinement with octree optimization, applied to a multi-LiDAR vehicle system. Recent advancements have further refined these approaches: Zhang \emph{et al.} \cite{b13} proposed PCR-CG, a point cloud registration method that explicitly incorporates color and geometric features, enhancing alignment accuracy in complex scenes, while Yu \emph{et al.} \cite{b14} introduced PEAL, a prior-embedded explicit attention learning framework tailored for low-overlap point cloud registration, addressing challenges in sparse overlap scenarios.

In addition, deep learning-based calibration methods have gained traction. For instance, PointNet++ \cite{b31} extracts robust point cloud features for registration, while DeepVCP \cite{b32} leverages deep neural networks to predict correspondences for LiDAR calibration. These methods often achieve high accuracy but typically require overlapping FoVs, extensive training data, and significant computational resources, limiting their applicability to non-overlapping configurations or resource-constrained platforms. Although these approaches offer flexibility, they typically require overlapping fields of view and remain sensitive to noise and outliers.

\subsection{Motion-Based Calibration}
Motion-based methods exploit the relative motion of LiDAR sensors on a moving platform to estimate extrinsic parameters. These techniques may integrate external sensors, such as GNSS or IMUs \cite{b15}, or depend solely on LiDAR data \cite{b16}. Common implementations use Visual-Inertial Odometry (VIO) \cite{b17} or LiDAR-Inertial Odometry (LIO) \cite{b18} to reconstruct sensor trajectories and compute transformations. For instance, Chang \emph{et al.} \cite{b19} proposed a graph optimization framework to jointly refine extrinsic parameters and trajectories. While these methods do not necessitate overlapping fields of view, they are prone to drift over extended periods due to cumulative odometry errors.

Several studies closely related to our work focus on target-free calibration using ICP to align natural scene point clouds. De Miguel \emph{et al.} \cite{b20} employed ICP to match the accumulated global point cloud of a target LiDAR with frame-by-frame clouds from a source LiDAR, iteratively tightening the matching threshold to tolerate initial errors. However, this approach overlooks errors in the accumulated point cloud, potentially compromising accuracy. In contrast, Liu \emph{et al.} \cite{b21} introduced a target-free method combining adaptive voxelization and Bundle Adjustment (BA) to optimize extrinsic parameters and poses across multi-view point clouds. Their voxelization strategy efficiently extracts planar features, enhancing robustness and reducing computational load. Yet, this method falters in non-overlapping field-of-view scenarios due to its dependence on geometric consistency, and its performance degrades with sparse features or large initial errors.

Current multi-LiDAR calibration methods face notable limitations. Most rely on overlapping fields of view, restricting their use in narrow or non-overlapping configurations. Moreover, their dependence on precise initial extrinsic estimates increases complexity and limits flexibility in dynamic or unknown settings, hindering broader adoption in practice. To address these issues, we propose a target-free calibration method that eliminates the need for overlapping fields of view and tolerates initial errors up to 0.4 m and 30°, significantly improving robustness and applicability.

\section{Methodology}
\label{sec:method}

\subsection{Notation and Symbols}
\label{sec:notation}
To enhance clarity, we summarize the primary symbols used in this paper in Table \ref{tab:symbols}. These notations are consistently applied throughout the methodology and experiments.

\begin{table}[!t]
\caption{Primary Symbols and Notations Used in the Paper}
\label{tab:symbols}
\centering
\setlength{\tabcolsep}{3pt}
\resizebox{\columnwidth}{!}{%
\begin{tabular}{|c|l|}
\hline
Symbol & Description \\
\hline
\( L_A^{t_j} \) & Point cloud data from LiDAR A at time \( t_j \) \\
\( L_B^{t_j} \) & Point cloud data from LiDAR B at time \( t_j \) \\
\( C_A^{map} \) & Global reference point cloud map from LiDAR A \\
\( T_A^{t_j} \) & Pose of LiDAR A at time \( t_j \), \( T_A^{t_j} \in SE(3) \) \\
\( T_B^A \) & Extrinsic transformation from LiDAR B to LiDAR A, \( T_B^A = (R_B^A, t_B^A) \) \\
\( R_B^A \) & Rotation matrix from LiDAR B to LiDAR A, \( R_B^A \in SO(3) \) \\
\( t_B^A \) & Translation vector from LiDAR B to LiDAR A, \( t_B^A \in \mathbb{R}^3 \) \\
\( \xi \) & Lie algebra parameter for \( T_B^A \), \( \xi \in \mathfrak{se}(3) \) \\
\( (\cdot)^\wedge \) & Operator mapping a vector to a skew-symmetric matrix \\
\( n_v \) & Normal vector of the \( v \)-th voxel plane \\
\( c_v \) & Centroid of the \( v \)-th voxel plane \\
\( \omega_v \) & Confidence weight of the \( v \)-th voxel \\
\( N_v \) & Number of points in the \( v \)-th voxel for optimization \\
\( \delta \) & Convergence threshold for iterative optimization \\
\( N \) & Maximum number of iterations \\
\( \eta \) & Planarity index, \( \eta = \frac{\lambda_1}{\lambda_2 + \lambda_3} \) \\
\( \lambda_1, \lambda_2, \lambda_3 \) & Eigenvalues of voxel covariance matrix \\
\( \text{roll}, \text{pitch}, \text{yaw} \) & Euler angles in ZYX Tait-Bryan convention \\
\hline
\end{tabular}
}
\end{table}

\subsection{Overview}
\label{sec:overview}
\begin{figure*}[!t]
    \centerline{\includegraphics[width=\textwidth]{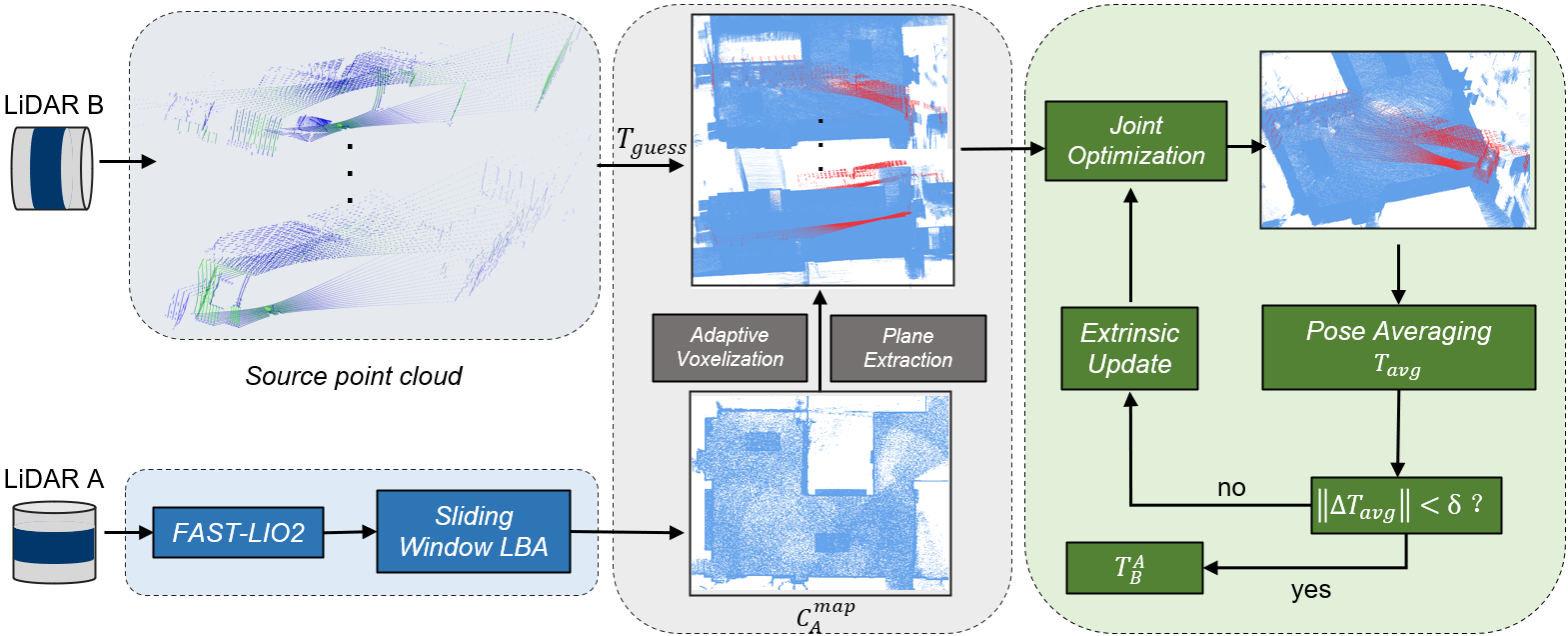}}
    \caption{System overview. The target LiDAR A accumulates an accurate point cloud map \( C_A^{map} \) using sliding-window LBA, enabling precise calibration without overlapping FoVs by aligning LiDAR B’s point clouds to the map. The map undergoes adaptive voxelization and feature extraction, followed by iterative matching with LiDAR B, optimizing the extrinsic parameters \( T_B^A \).}
    \label{fig:overview}
\end{figure*}

As shown in Fig. \ref{fig:overview}, our method achieves high-precision extrinsic calibration through three key steps: sliding-window LBA-optimized LiDAR odometry, adaptive voxel-based planar feature extraction, and parallel optimization of dual LiDAR extrinsics. LiDAR A, based on the initial scans and poses provided by the front-end odometry, performs local LBA optimization within a fixed-size sliding window to optimize its pose trajectory. This process incorporates scene constraints to enhance accuracy and reduce accumulated drift, thereby refining the local map. A dynamic voxelization strategy is then employed to extract high-quality planar features from LiDAR A’s accumulated point cloud, improving both the efficiency and adaptability of feature extraction. Finally, by jointly utilizing the accumulated point cloud map from LiDAR A and the multi-frame point cloud data from LiDAR B, we iteratively construct the reprojection error and optimize the extrinsics between the two LiDARs, denoted as \( T_B^A = (R_B^A, t_B^A) \in SE(3) \), where \( R_B^A \in SO(3) \) and \( t_B^A \in \mathbb{R}^3 \) represent the rotation and translation, respectively.

To intuitively explain how precise calibration is achieved without overlapping FoVs,one forward-facing (LiDAR A) and one rear-facing (LiDAR B), with non-overlapping FoVs. As the vehicle moves through an environment, LiDAR A continuously scans to build a global reference map \( C_A^{map} \) using optimized poses from sliding-window LBA. Although LiDAR B’s point clouds \( L_B^{t_j} \) do not overlap with LiDAR A’s in real-time, they capture parts of the same environment at different times. By aligning \( L_B^{t_j} \) to \( C_A^{map} \) through point-to-plane matching , we establish correspondences between the two LiDARs’ data, enabling the estimation of \( T_B^A \). This frame-to-map alignment replaces the need for direct FoV overlap, as the global map provides a consistent reference for calibration.

\subsection{Motion Compensation}
\label{sec:motion_comp}
During LiDAR scanning, the motion of the mobile platform introduces distortion in the point cloud, as points captured at different times within a single scan (one full rotation) correspond to different platform poses. To address this, we employ motion compensation, also known as point cloud deskewing, using the inertial measurement unit (IMU) data integrated by FastLIO2 \cite{b22}. For each LiDAR frame \( L_A^{t_j} \) or \( L_B^{t_j} \), FastLIO2 predicts the platform’s motion during the scan using IMU measurements and interpolates the pose for each point based on its timestamp. The point cloud is then transformed to a common reference time (e.g., the start of the scan) to produce a deskewed point cloud, ensuring geometric consistency. This process is critical for accurate pose estimation and point cloud registration, especially in dynamic environments. The deskewed point clouds are used as input for subsequent LBA optimization and extrinsic calibration.

\subsection{LiDAR Odometry}
\label{sec:lidar_odometry}
In the non-overlapping dual-LiDAR extrinsic calibration framework, LiDAR odometry estimates the motion trajectory of the target LiDAR using its point cloud data while optimizing the pose trajectory to reduce cumulative errors. It serves as the foundation for subsequent extrinsic calibration. The primary objective is to construct a high-quality global reference map \( C_A^{map} \) while ensuring global trajectory consistency and accuracy, providing a reliable input for planar feature extraction.

We employ FastLIO2 \cite{b22} as the front-end odometry, utilizing deskewed point cloud data \( L_A^{t_j} \) collected by LiDAR A at time \( t_j \) (where \( j = 1, 2, \ldots, n \), and \( n \) is the total number of frames) along with IMU data to generate the initial pose sequence \( \{T_A^{t_j}\} \). FastLIO2 integrates IMU measurements to predict platform motion and aligns point clouds using a tightly-coupled LiDAR-inertial odometry framework, providing high-precision pose estimation in local regions. However, long-term operation may lead to accumulated errors due to IMU drift or environmental noise. Therefore, sliding-window LBA is applied to optimize the pose trajectory.

The sliding window approach is applied to \( n \) frames of LiDAR A's point cloud data \( \{L_A^{t_j}\} \) and initial pose sequence \( \{T_A^{t_j}\} \). The parameters are defined as follows:  
- Window size: \( w \)  
- Step size: \( d \)  
- Number of overlapping frames: \( o = d/2 \). Thus, the sequence is divided into \( k = \lceil(n-d)/(d-o)\rceil +1 \) windows. For the \( m \)-th window (\( m=1,2,…,k \)), the frame range is  
\(
[t_{m,1},t_{m,w}]
\)  
,where  
\(
t_{m,1} = t_{1+(m-1)(d-o)}
\)  . 

In each window, the pose of the first frame \( T_A^{t_{m,1}} \) is kept fixed as the reference pose. We optimize the relative pose transformations of the subsequent frames, specifically the relative pose transformation between frame \( j \) and frame \( j+1 \), denoted as \( \Delta T_{t_{m,j}}^{t_{m,j+1}} \) (\( j = 1, 2, \dots, w-1 \)). By transforming the point clouds of frames \( j \) and \( j+1 \) into the same coordinate system, the objective is to minimize the point cloud registration error. The optimization function is formulated as:
\begin{equation}
\begin{split}
    \Delta T_{t_{m,j}}^{t_{m,j+1}} = \arg\min_{\Delta T} 
    \text{dist} \Big(
    T_A^{t_{m,j}} \cdot L_A^{t_{m,j}}, \\
    (T_A^{t_{m,j}} \cdot \Delta T_{t_{m,j}}^{t_{m,j+1}}) \cdot L_A^{t_{m,j+1}}
    \Big),
\end{split}
\label{eq:odometry}
\end{equation}
where \( \text{dist}(\cdot, \cdot) \) represents the point-to-plane registration error. After optimization, the updated pose of the \( j \)-th frame is given by:
\begin{equation}
T_A^{t_{m,j}^\ast} = T_A^{t_{m,1}} \cdot \prod_{k=1}^{j-1} \Delta T_{t_{m,j}}^{t_{m,j+1}}.
\label{eq:pose_update}
\end{equation}

To ensure the smoothness of the global trajectory, the overlapping region between adjacent windows is utilized. The optimized poses of the last \( o \) frames from the \((m-1)\)-th window, \( \{T_A^{t_{m-1,w-o+1}^\ast}, \dots, T_A^{t_{m-1,w}^\ast} \} \), are used as the initial values for the first \( o \) frames of the \( m \)-th window, with additional constraints incorporated into the optimization.

The optimization process adopts the BALM \cite{b23} method, leveraging multi-threaded parallel computation of the Hessian matrix and Jacobian vectors for each window to improve computational efficiency. The Hessian matrix for each window has dimensions \( 6(w-1) \times 6(w-1) \), where \( w \) is the window size and 6 represents the degrees of freedom in \( SE(3) \). The Jacobian vectors are computed for each point-to-plane constraint, with dimensions proportional to the number of points in the window (typically sparse due to planar feature selection). This sparsity and parallel computation reduce the computational complexity from \( O(n^3) \) to approximately \( O(n) \) per iteration for large point clouds, where \( n \) is the number of points \cite{b27}. Finally, the optimized poses across all windows are smoothly merged into a global sequence \( \{T_A^{t_j\ast} \}_{j=1,2,\dots,n} \), which is used to construct the reference map \( C_A^{map} \):
\begin{equation}
C_A^{map} = \bigcup_{j} \left( T_A^{t_j\ast} \cdot L_A^{t_j} \right).
\label{eq:map}
\end{equation}

\begin{figure}[!t]
    \centerline{\includegraphics[width=\columnwidth]{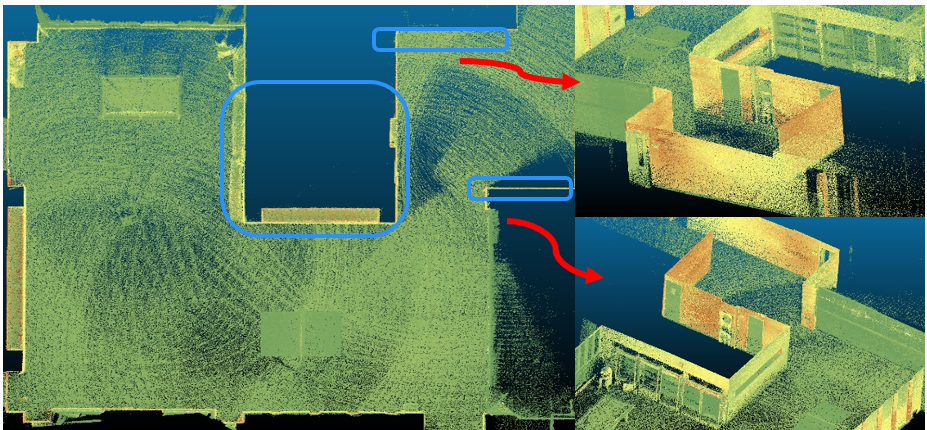}}
    \caption{Scene used to compare LBA effect, with planar features (blue regions) critical for alignment. Top right: pre-optimized point cloud; bottom right: post-optimized point cloud. Red arrows indicate alignment differences. Slight FoV variations result from pose adjustments during LBA.}
    \label{fig:scene}
\end{figure}

\begin{figure}[!t]
    \centerline{\includegraphics[width=1.05\columnwidth]{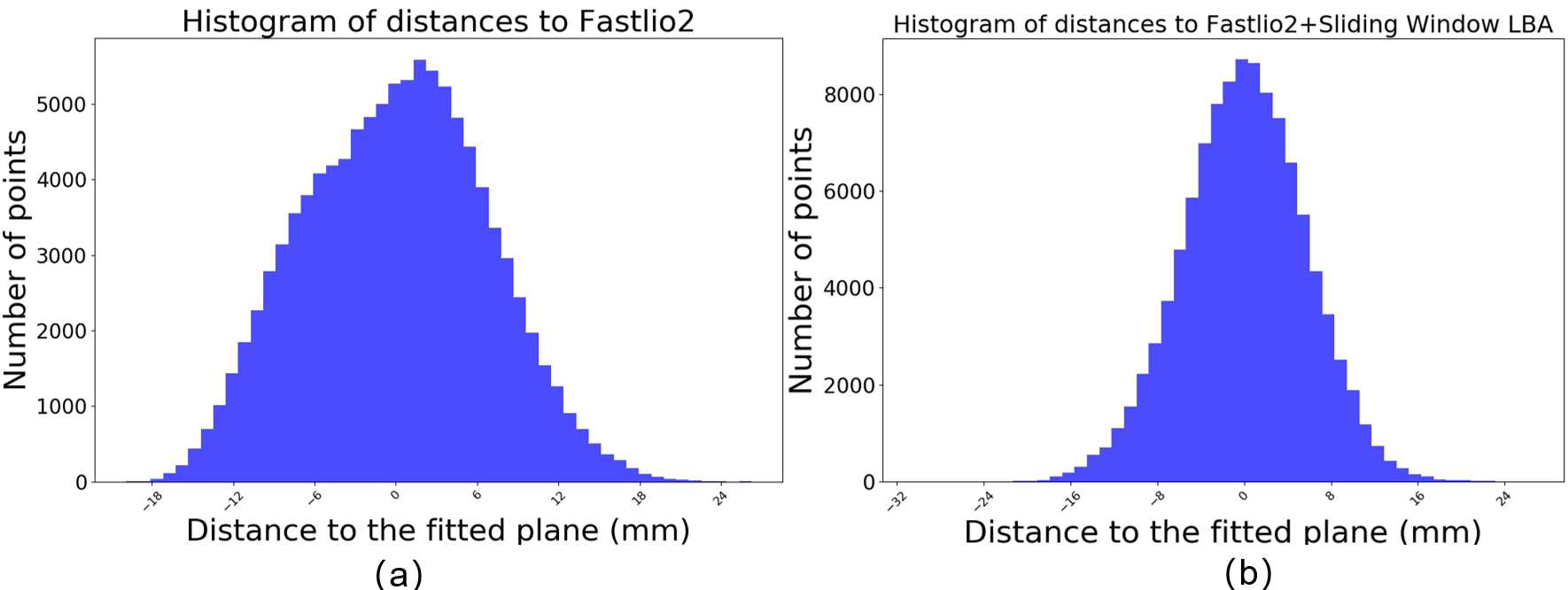}}
    \caption{Distribution of points fitted to the point cloud plane before and after LBA optimization. Post-optimization points (b) are more concentrated around zero, indicating a closer fit to the true plane compared to pre-optimization (a).}
    \label{fig:distribution}
\end{figure}

To demonstrate the effectiveness of LBA optimization, we present scene reconstruction results in Fig. \ref{fig:scene}. The figure depicts a test environment with planar features (blue regions) essential for point-to-plane alignment. Red arrows highlight misalignment in the pre-optimized point cloud (top right) compared to enhanced alignment in the post-optimized point cloud (bottom right). The subtle field-of-view (FoV) differences between the two point clouds stem from pose refinements during LBA, which improves the global map’s fidelity to the true environment. This is further supported by Fig. \ref{fig:distribution}, which shows the distribution of outlier errors for fitted planes, clearly indicating that the plane after LBA optimization is significantly closer to the true geometry.

\subsection{Adaptive Map Voxelization and Plane Extraction}
\label{sec:voxelization}
The previous section optimized the pose trajectory of the target LiDAR (LiDAR A) through sliding-window LBA, constructing a high-quality global reference map \( C_A^{map} \). To further enhance the accuracy and robustness of extrinsic calibration, this section focuses on extracting high-quality planar features from \( C_A^{map} \), introducing an adaptive voxelization method, and optimizing the planar extraction process.

We adopt an adaptive voxelization strategy \cite{b24}, where \( C_A^{map} \) is first uniformly divided into large initial voxels (e.g., initial voxel size \( l_{\text{parent}} = 1 \, \text{m} \)) to quickly cover the entire scene. Each voxel is then evaluated for planarity, and if the geometric features are complex, the voxel is further subdivided into smaller sub-voxels. For each voxel, we compute the covariance matrix \( C \) of its point cloud and perform eigenvalue decomposition:
\begin{align}
C &= \frac{1}{M} \sum_{i=1}^{M} (p_i - \bar{p})(p_i - \bar{p})^T \notag \\
&= U \Lambda U^T, \quad \Lambda = \operatorname{diag}(\lambda_1, \lambda_2, \lambda_3).
\label{eq:voxel}
\end{align}

  A planarity index is defined as \( \eta = \frac{\lambda_1}{\lambda_2 + \lambda_3} \). If \( \eta < \tau_\eta \) (e.g., \( \tau_\eta = 0.1 \)), the voxel is classified as a planar voxel; otherwise, it is subdivided into eight sub-voxels, each with edge length \( l_{\text{child}} = \frac{l_{\text{parent}}}{2} \). This process continues recursively until a termination condition is met (e.g., maximum recursion depth or planarity threshold satisfied).

To enhance the quality of planar features and minimize redundancy, we implement a neighboring voxel merging process. For adjacent voxels, we compute the angle \( \theta \) between their normal vectors \( \mathbf{n}_i \) and \( \mathbf{n}_j \), defined as \( \theta = \cos^{-1} (\mathbf{n}_i^T \mathbf{n}_j) \), and the Euclidean distance between their centroids \( q_i \) and \( q_j \), given by \( d = \| q_i - q_j \| \). If both \( \theta < \tau_\theta \) and \( d < \tau_d \), where \( \tau_\theta \) and \( \tau_d \) are predefined thresholds, the voxels are merged to reduce redundant computations.

For each resulting voxel, we extract its planar parameters \( (\mathbf{n}, q) \), where \( \mathbf{n} \) is the unit normal vector and \( q \) is the centroid, and compute a confidence weight as follows:
\begin{equation}
\omega_v = \frac{M_v}{1 + \sigma_\lambda} \cdot e^{-\gamma \eta},
\label{eq:weight}
\end{equation}
where \( M_v \) is the number of points within the voxel, \( \sigma_\lambda \) is the eigenvalue standard deviation, and \( \gamma \) is an attenuation coefficient. This weighting factor is used in the subsequent weighted optimization for extrinsic calibration, enhancing robustness.

\subsection{Dual-LiDAR Extrinsic Calibration}
\label{sec:dual_lidar}
After completing adaptive voxelization and planar feature extraction based on \( C_A^{map} \), this section aligns the multi-frame point cloud data from the source LiDAR (LiDAR B) with the target reference map, constructs the point-to-plane reprojection error, and jointly optimizes the extrinsic parameters between the two LiDARs. The initial extrinsic \( T_{guess} \) can be roughly estimated through manual measurement or prior knowledge of the installation configuration. The method is tolerant to initial errors, allowing a translation error of up to 40 cm and a rotation error of up to 30°, ensuring convergence without requiring precise initial values.

The optimized target reference map is denoted as \( C_A^{map} \), and the point cloud data collected by the source LiDAR is represented as \( L_B^{t_j} = \{L_B^{t_1}, L_B^{t_2}, \dots, L_B^{t_n} \} \), which is one-to-one corresponding and synchronized with the target LiDAR’s pose sequence \( T_A^{t_j}, j \in [1, n] \) and the accumulated \( C_A^{map} \) within the same time period.

Through voxel construction and planar feature extraction, each voxel contains a set of points that form a plane, thereby establishing planar constraints between the scanned frame point clouds and the target map. Specifically, for the \( v \)-th voxel, its planar features are represented by the normal vector \( n_v \in \mathbb{R}^3 \) and the centroid \( c_v \in \mathbb{R}^3 \). The distance from a source LiDAR point \( p_i^B \in \mathbb{R}^3 \) to the target voxel plane is defined as the residual:
\begin{equation}
r_i = n_v^{\top} (R_B^A p_i^B + t_B^A - c_v).
\label{eq:residual}
\end{equation}

The global objective function is formulated as the weighted sum of squared residuals:
\begin{equation}
L_{\text{Global}} = \sum_{v \in V} \omega_v \sum_{i=1}^{N_v} \| r_i \|^2,
\label{eq:objective}
\end{equation}
where \( \omega_v \) is the confidence weight, and \( N_v \) represents the number of points used for optimization in the \( v \)-th voxel.

To facilitate optimization, the extrinsic transformation \( T_B^A \in SE(3) \) is parameterized using the Lie algebra \( \mathfrak{se}(3) \), with the parameter vector \( \xi \in \mathbb{R}^6 \). The operator \( (\cdot)^\wedge \) maps a 6D vector to a 4×4 skew-symmetric matrix, as defined in \cite{b26}. The Jacobian of the residual with respect to \( \xi \) is computed as:
\begin{equation}
\frac{\partial r_i}{\partial \xi} = - n_v^\top \begin{bmatrix} R_B^A (p_i^B)^\wedge & I_3 \end{bmatrix},
\label{eq:jacobian}
\end{equation}
where \( (p_i^B)^\wedge \) is the skew-symmetric matrix of \( p_i^B \), and \( I_3 \) is the 3×3 identity matrix.

Using multi-threaded parallel computation, the Hessian matrix is constructed as \( H = J^\top J + \mu I \), where \( \mu \) is a damping factor. The Levenberg-Marquardt (LM) algorithm \cite{b27} solves for the increment \( \Delta \xi = -H^{-1} J^\top r \). The extrinsic parameters are iteratively updated as:
\begin{equation}
T_B^A \gets T_B^A \cdot \exp{(\Delta \xi^\wedge)},
\label{eq:update}
\end{equation}
the process is iterated until the change in the external parameters is smaller than the convergence threshold $\delta$, or the maximum number of iterations $N$ is reached. The iterative process is summarized in Algorithm 1.
\begin{algorithm}[!h]
    \caption{Robust Extrinsic Calibration of Non-Overlapping LiDARs}
    \label{alg:AOA}
    \renewcommand{\algorithmicrequire}{\textbf{Input:}}
    \renewcommand{\algorithmicensure}{\textbf{Output:}}
    \begin{algorithmic}[1]
        \REQUIRE $C_A^{map},\, L_B^{t_j}\ (j=1,\dots,n),\, T_{guess},\, N,\, \delta$
        \ENSURE $T_B^A$
        \STATE $i \leftarrow 0$, $T_{prev} \leftarrow T_{guess}$
        \WHILE{$i < N$}
            \FOR{$t \leftarrow 1$ \TO $n$}
                \STATE $T_{local}^t \leftarrow \text{BA\_Optimize}(C_A^{map},\, L_B^{t_j},\, T_{prev})$ \\
                \quad \text{(Equations (6)-(9)}
            \ENDFOR
            \STATE $T_{avg} \leftarrow \text{Average}(\{T_{local}^t\}_{t=0}^{t=n})$
            \IF{$\|T_{avg} - T_{prev}\| < \delta$}
                \STATE \textbf{break}
            \ENDIF
            \STATE $T_{prev} \leftarrow T_{avg}$
            \STATE $i \leftarrow i + 1$
        \ENDWHILE
        \STATE $T_B^A \leftarrow T_{avg} \cdot T_{guess}$
        \RETURN $T_B^A$
    \end{algorithmic}
\end{algorithm}
\label{sec:experiments}

\subsection{Error Metrics}
\label{sec:error_metrics}
To evaluate the accuracy of our calibration method, we use translation and rotation errors. Let \( T_{\text{est}} = (R_{\text{est}}, t_{\text{est}}) \) be the estimated extrinsic transformation, and \( T_{\text{gt}} = (R_{\text{gt}}, t_{\text{gt}}) \) be the ground truth. The translation error is the Euclidean distance:
\begin{equation}
e_{\text{trans}} = \| t_{\text{est}} - t_{\text{gt}} \|_2,
\label{eq:trans_error}
\end{equation}
measured in meters (m). The rotation error is the angular difference:
\begin{equation}
e_{\text{rot}} = \arccos\left( \frac{\text{trace}(R_{\text{gt}}^T R_{\text{est}}) - 1}{2} \right),
\label{eq:rot_error}
\end{equation}
measured in radians (rad). These metrics are standard in robotics \cite{b30}.

\subsection{Experimental Setup}
\begin{figure}[!t]
    \centerline{\includegraphics[width=0.5\columnwidth]{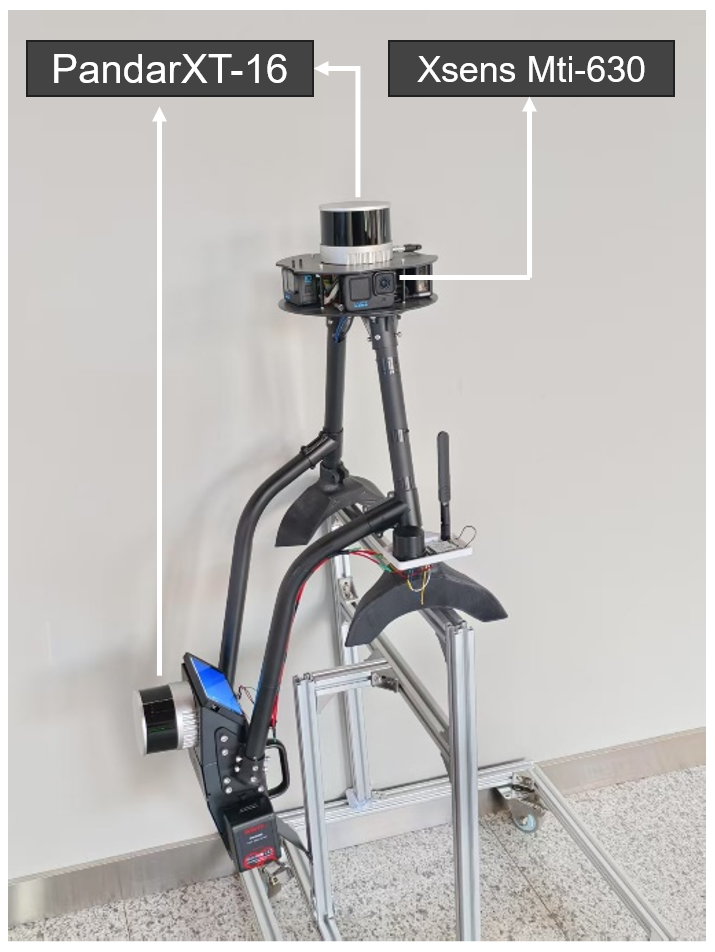}}
    \caption{Backpack-mounted data collection platform with two PandarXT16 LiDARs and an Xsens Mti-630 IMU.}
    \label{fig:backpack}
\end{figure}

This section outlines the experimental setup to evaluate the robustness and accuracy of our calibration method, leveraging both simulation and real-world environments. In simulation, we use CARLA to collect data under varied configurations, obtaining ground truth extrinsic parameters for precise accuracy assessment. In the real-world setup, we employ a backpack-mounted platform (Fig. \ref{fig:backpack}), equipped with two PandarXT16 LiDARs and an Xsens Mti-630 IMU. One LiDAR is mounted horizontally atop the device, the other vertically on the back, with the IMU fixed below the horizontal LiDAR. This portable system enables flexible data collection across diverse environments. The LiDARs exhibit minimal overlapping fields of view, posing unique calibration challenges. Intrinsic parameters of both LiDARs are pre-calibrated by the manufacturer. Our method is referred to as DLBAcalib in the following description.

\subsection{Simulation Environment Setup}
\label{sec:sim_setup}
We leverage the CARLA simulation environment \cite{b25} to rigorously evaluate our method. CARLA provides a versatile platform for simulating LiDAR sensors with configurable parameters such as beam patterns, measurement noise, and mounting positions. We simulate a Velodyne-16-line LiDAR with a vertical field of view of 30° and a measurement noise characterized by a standard deviation of 0.01 m for point distances. Motion distortion is modeled during each LiDAR scan,  with points timestamped and adjusted based on interpolated poses from simulated IMU data. Point clouds are deskewed as described in Section \ref{sec:motion_comp}. Multiple datasets are acquired, each comprising point clouds from two LiDARs, their relative poses, trajectories, timestamps, and synchronized IMU data.

We conducted experiments with different external parameters between two LiDARs, as listed in Table \ref{tab:config}. Configuration 1 matches our backpack platform. The roll, pitch, and yaw angles follow the ZYX Tait-Bryan convention \cite{b29}, with rotations applied first around the Z-axis (yaw), then Y-axis (pitch), and finally X-axis (roll).

\begin{table}[!t]
\caption{Configurations of Dual LiDARs in the CARLA Simulation Environment}
\label{tab:config}
\centering
\setlength{\tabcolsep}{3pt}
\resizebox{\columnwidth}{!}{%
\begin{tabular}{|c|c|c|c|c|c|c|}
\hline
Config & X (m) & Y (m) & Z (m) & Roll (°) & Pitch (°) & Yaw (°) \\
\hline
1 & 0 & -0.35 & -0.9 & -90 & 0 & 180 \\
2 & 0 & 0.5 & 0.5 & 90 & 0 & 90 \\
3 & -1 & 0 & 0 & 180 & 0 & 202 \\
4 & 0.6 & 0.4 & 0.4 & 0 & 90 & 180 \\
5 & 0.4 & -0.2 & -1 & 180 & -90 & 0 \\
\hline
\end{tabular}
}
\end{table}

\subsection{Simulation Experiment Results}
\label{sec:sim_results}
\begin{figure}[!t]
    \centering
    \includegraphics[width=1\columnwidth]{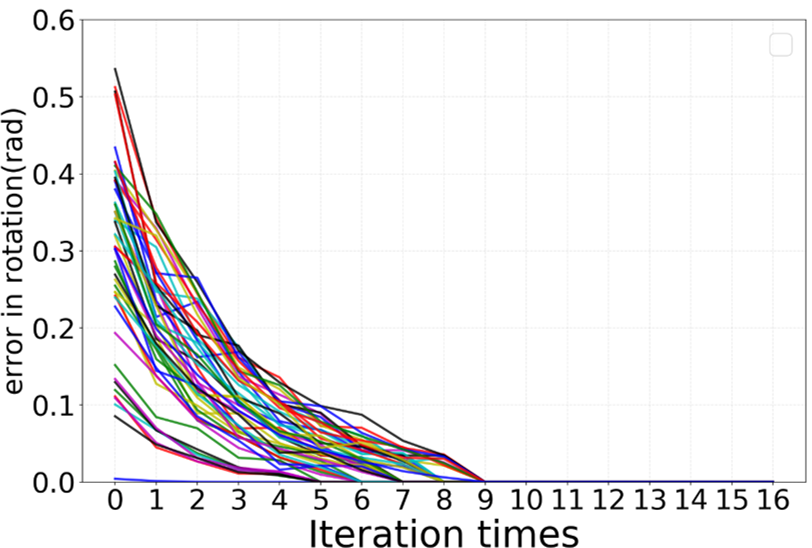}
    \vspace{0.25em} 
    \includegraphics[width=1\columnwidth]{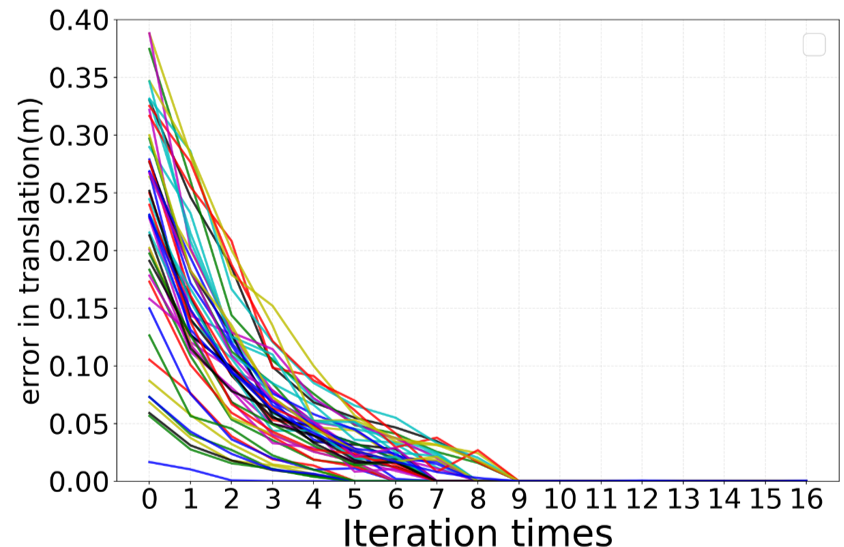}
    \caption{Variation of translation and rotation errors with iterations for 10 representative experiments under different initial conditions (±0.4 m translation, ±30° rotation).}
    \label{fig:iteration}
\end{figure}

To evaluate the robustness to initial conditions, we conducted 50 experiments under each LiDAR configuration, with variations in motion trajectory and initial extrinsic parameters. Initial estimates were obtained by randomly adding errors to the true calibration values, with translation errors up to ±0.4 m and rotation errors up to ±30°. The same random seed was used for fair comparison across methods. Table \ref{tab:avg_error} summarizes the average translation and rotation errors across 10, 20, and 50 trials for Configuration 1, demonstrating consistent performance with increasing trial counts. Fig. \ref{fig:iteration} displays 10 representative experiments, showing that DLBAcalib converges to smaller errors under various initial perturbations.

\begin{figure}[!t]
    \centering
    \includegraphics[width=1\columnwidth]{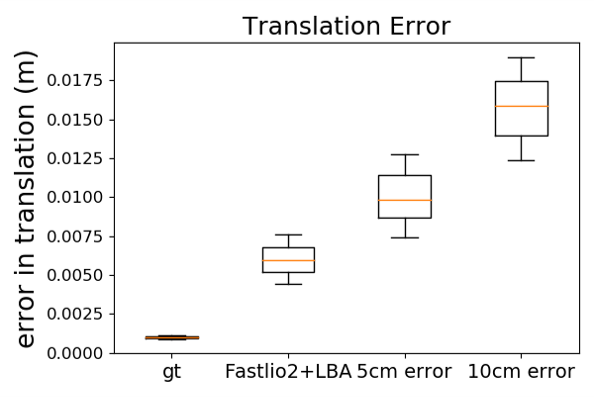}
    \vspace{0.25em}
    \includegraphics[width=1\columnwidth]{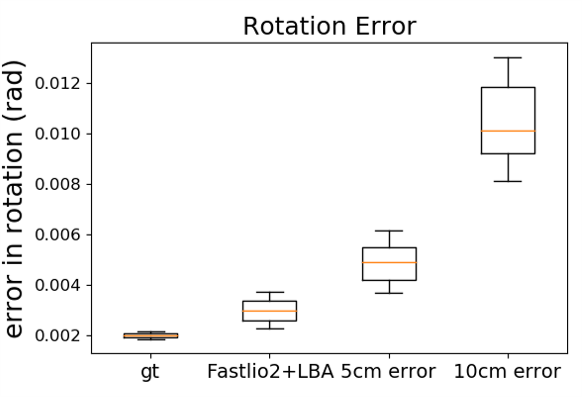}
    \caption{Comparison of results with different front-end LiDAR trajectory errors. From left to right: ground-truth trajectory, FastLIO2+LBA optimized trajectory, ground-truth with 5 cm error, and ground-truth with 10 cm error.}
    \label{fig:odometry_error}
\end{figure}

The accuracy of front-end odometry significantly influences calibration performance. We tested varying localization accuracies: ground truth trajectories, degraded trajectories with Gaussian noise, and FastLIO2 outputs, as shown in Fig. \ref{fig:odometry_error}. Results indicate robustness to localization inaccuracies, with our LBA minimizing errors effectively.

We compared our method against MLCC \cite{b21} and a GICP-based method \cite{b20}, using identical input data: FastLIO2 pose estimates and \( C_A^{map} \). Table \ref{tab:comparison} shows that DLBAcalib outperforms both in non-overlapping configurations, particularly with large initial errors or sparse features. Compared to recent deep learning-based methods like DeepVCP \cite{b32}, our approach does not require training data or overlapping FoVs, making it more suitable for non-overlapping scenarios, though deep learning methods may excel in overlapping FoV cases with sufficient training data.

\begin{table}[!t]
\caption{Average Translation and Rotation Errors for Configuration 1 Across Multiple Trials}
\label{tab:avg_error}
\centering
\setlength{\tabcolsep}{3pt}
\resizebox{\columnwidth}{!}{%
\begin{tabular}{|c|c|c|}
\hline
Trials & Avg. Translation Error (m) & Avg. Rotation Error (rad) \\
\hline
10 & 0.0052 & 0.0041 \\
20 & 0.0051 & 0.0040 \\
50 & 0.0050 & 0.0041 \\
\hline
\end{tabular}
}
\end{table}
\begin{table}[!t]
\caption{Comparison of Translation Error (m) and Rotation Error (rad) for Non-Overlapping Configurations 1--5 of Table \ref{tab:config} between DLBAcalib, MLCC \cite{b21}, and PCDcalib \cite{b20}}
\label{tab:comparison}
\centering
\setlength{\tabcolsep}{2pt}
\resizebox{\columnwidth}{!}{%
\begin{tabular}{|c|l|c|c|}
\hline
Config & Method & Translation Error (m) & Rotation Error (rad) \\
\hline
1 & MLCC      & 0.00964 & 0.04933 \\
  & PCDcalib  & 0.02254 & 0.04362 \\
  & DLBAcalib & \textbf{0.00509} & \textbf{0.00409} \\
\hline
2 & MLCC      & 0.00669 & 0.00489 \\
  & PCDcalib  & 0.00985 & \textbf{0.00126} \\
  & DLBAcalib & \textbf{0.00578} & 0.00367 \\
\hline
3 & MLCC      & 0.01693 & 0.00599 \\
  & PCDcalib  & 0.01356 & 0.00465 \\
  & DLBAcalib & \textbf{0.00514} & \textbf{0.00336} \\
\hline
4 & MLCC      & 0.00951 & 0.00464 \\
  & PCDcalib  & 0.01189 & \textbf{0.00347} \\
  & DLBAcalib & \textbf{0.00580} & 0.00396 \\
\hline
5 & MLCC      & 0.00974 & 0.00633 \\
  & PCDcalib  & 0.01255 & \textbf{0.00196} \\
  & DLBAcalib & \textbf{0.00545} & 0.00418 \\
\hline
\end{tabular}
}
\end{table}



\subsection{Real-World Experiment Results}
\label{sec:real_results}
\begin{figure}[!t]
    \centerline{\includegraphics[width=\columnwidth]{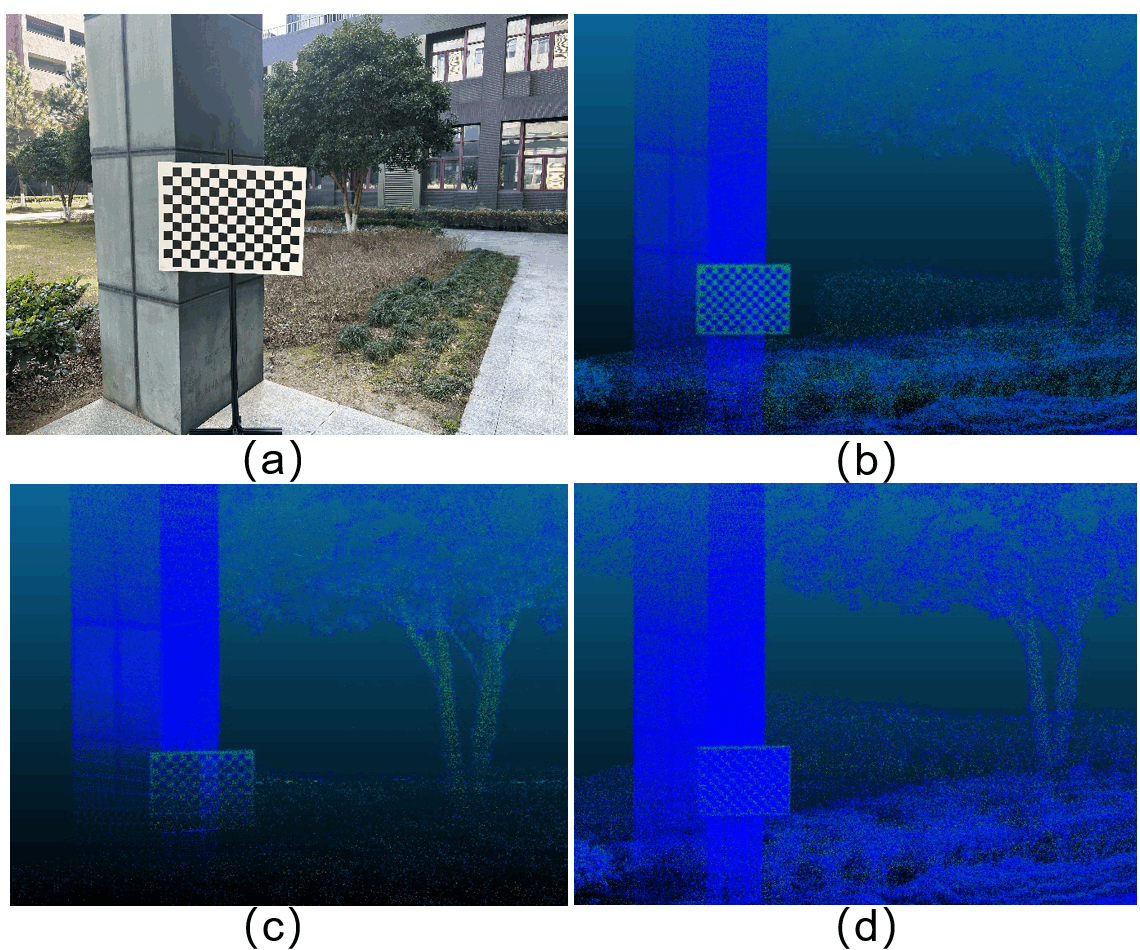}}
    \caption{Scenario 2: (a) Scene with outliers, (b) DLBAcalib, (c) MLCC, (d) PCDcalib results. DLBAcalib shows superior alignment.}
    \label{fig:scenario2}
\end{figure}

\begin{figure}[!t]
    \centerline{\includegraphics[width=\columnwidth]{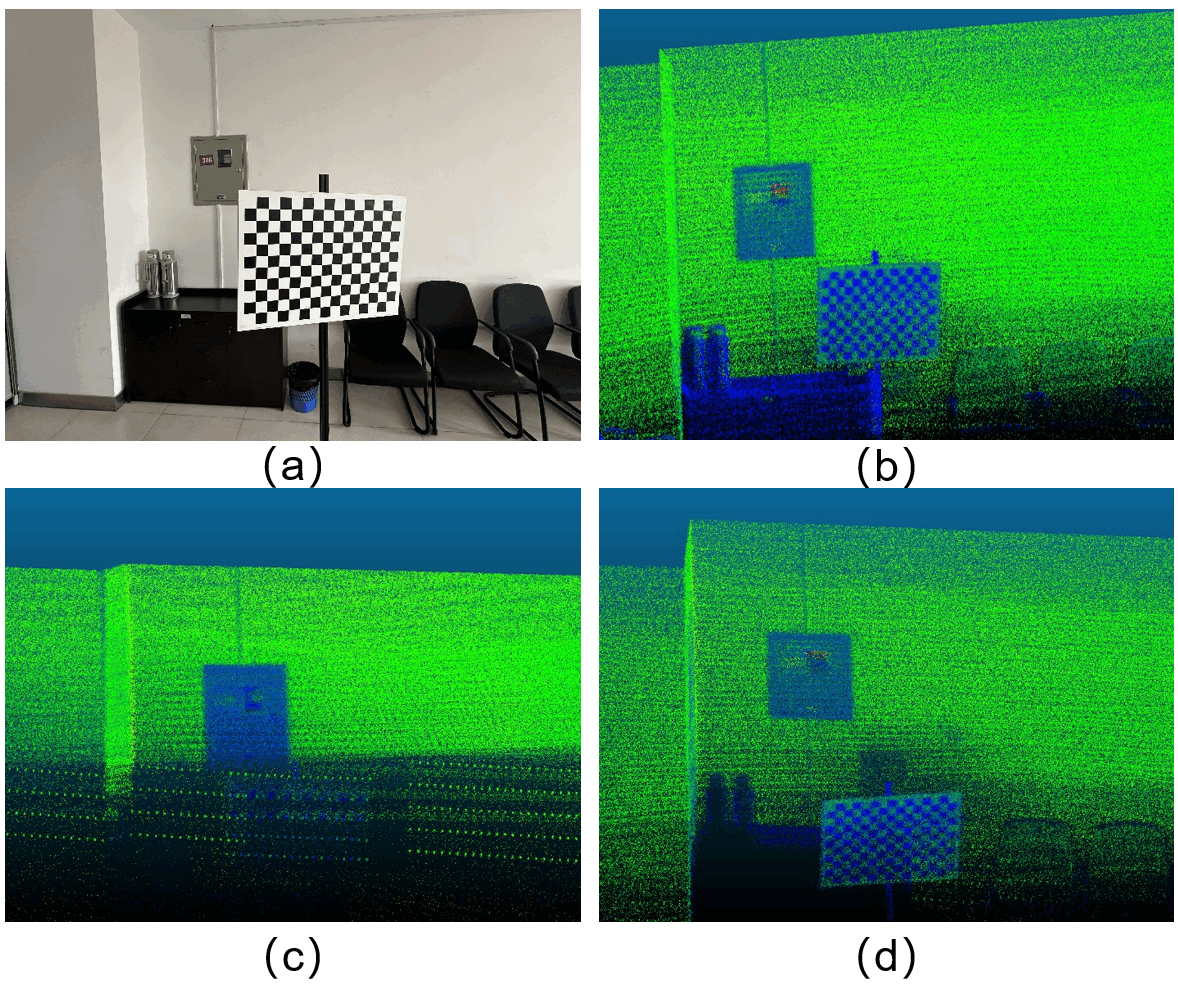}}
    \caption{Scenario 3: (a) Non-overlapping scene, (b) DLBAcalib, (c) MLCC, (d) PCDcalib results. DLBAcalib performs best in non-overlapping setups.}
    \label{fig:scenario3}
\end{figure}

Real-world experiments were conducted using the backpack platform. We selected three scenarios: 1) a structured environment with a calibration board and walls for high-precision reconstruction, 2) a scene with significant outliers to test robustness, and 3) a non-overlapping FoV configuration to validate applicability. Initial extrinsic parameters were perturbed with random errors up to ±0.4 m in translation and ±30° in rotation, matching the simulation setup. Algorithm 1 converged in an average of 8 iterations across all scenarios, demonstrating robust performance. Results are shown in Figs. \ref{fig:scenario2} and \ref{fig:scenario3}. DLBAcalib consistently outperformed MLCC and PCDcalib, enhancing scene representation and downstream perception.

\section{Conclusion}
\label{sec:conclusion}
We propose a novel targetless extrinsic calibration method for dual LiDARs, addressing non-overlapping fields of view and imprecise initial parameters. By combining sliding-window LiDAR bundle adjustment (LBA) with adaptive voxelization, we eliminate the need for markers or manual tuning. The method reduces cumulative errors by optimizing the target LiDAR's trajectory and extracting planar features. Experiments show 5\,mm translation and 0.2$^\circ$ rotation errors, with robustness to initial errors up to 0.4\,m and 30$^\circ$. Our approach outperforms current methods and is suitable for autonomous driving and 3D modeling. Future work will focus on improving robustness, efficiency, and extending the approach to dynamic environments and additional sensors.


\begin{IEEEbiography}[{\includegraphics[width=1in,height=1.25in,clip,keepaspectratio]{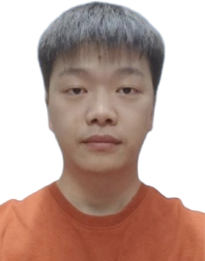}}]{Han Ye}
received the B.Eng. degree in automation from the Zhejiang University of Technology, Hangzhou, China, in 2023, where he is currently pursuing the master's degree in electronic information with a focus on control engineering at the College of Information Engineering.\\
\indent His current research interests include information fusion, computer vision, and localization.
\end{IEEEbiography}
\begin{IEEEbiography}[{\includegraphics[width=1in,height=1.25in,clip,keepaspectratio]{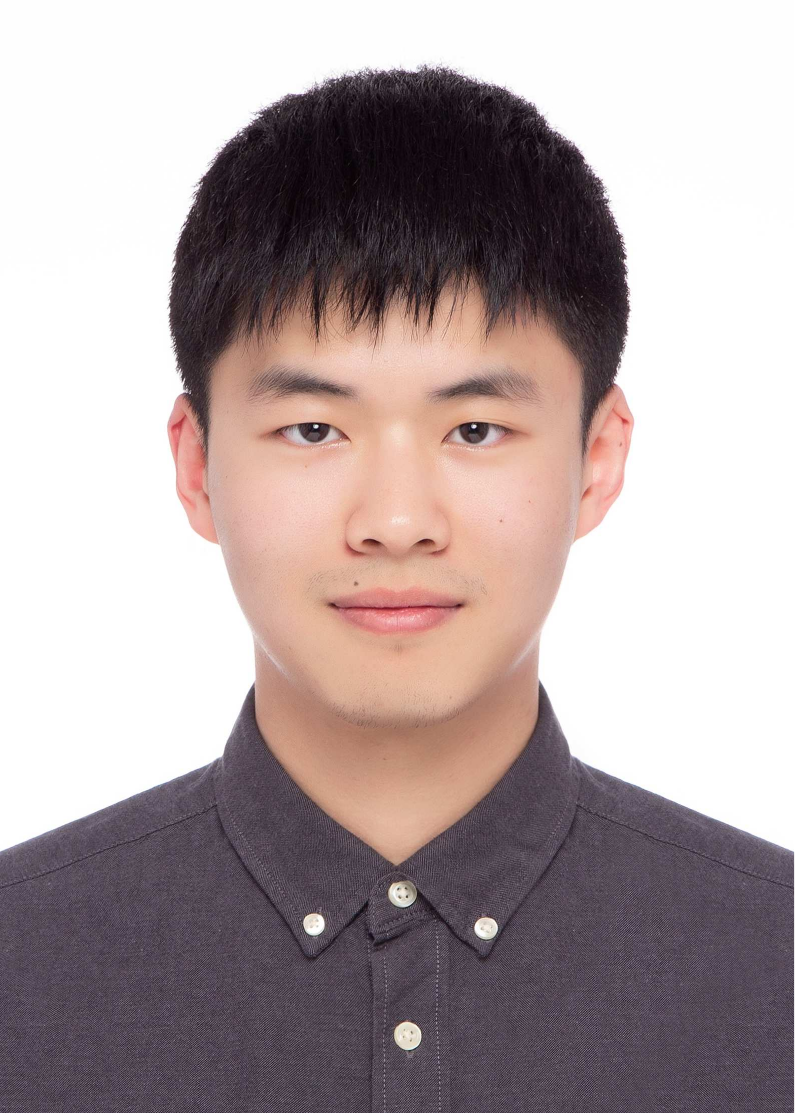}}]{Yuqiang Jin (Graduate Student Member, IEEE)} 
received the B.Eng. degree from the Department of Automation, Zhejiang University of Technology, Hangzhou, China, in 2020, where he is currently pursuing the Ph.D. degree in control science and engineering.\\
\indent His current research interests include information fusion, robotics vision, and localization and mapping for autonomous robots.
\end{IEEEbiography}
\begin{IEEEbiography}[{\includegraphics[width=1in,height=1.25in,clip,keepaspectratio]{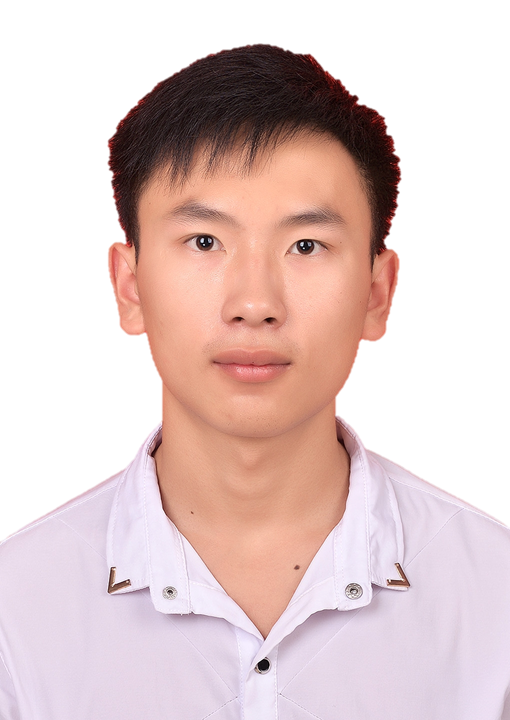}}]{Jinyuan Liu}
Ph.D. Student (2021.09-)
He is currently pursuing a Ph.D. degree with the College of Information Engineering, Zhejiang University of Technology.\\ \indent He received the B.S. degree from the Department of Physics, Zhejiang University of Technology, Hangzhou, China, in 2021, where he is currently pursuing the Ph.D. degree in control science and engineering.\\ \indent His research interests include modeling of autonomous robots, trajectory planning, and human-robot interaction.
\end{IEEEbiography}
\begin{IEEEbiography}[{\includegraphics[width=1in,height=1.25in,clip,keepaspectratio]{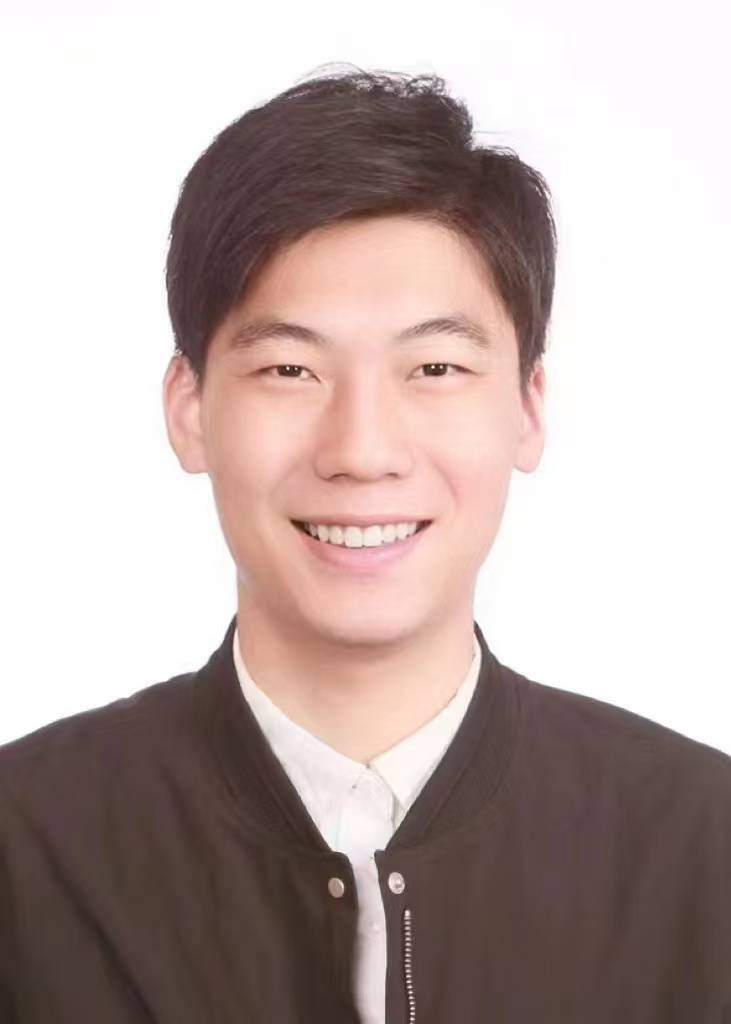}}]{Tao Li (Member, IEEE)}
received the B.S. degree in navigation engineering from Wuhan University,Wuhan, China, in 2018,and the Ph.D. degree in information and communication engineering from Shanghai Jiao Tong University, Shanghai, China,in 2023. He is currently an Associate Researcher with Zhejiang University of Technology.\\ \indent His current research interests include visual-SLAM, LiDARSLAM, global navigation satellite systems (GNSS),inertial navigation systems (INS), and information fusion.
\end{IEEEbiography}
\begin{IEEEbiography}[{\includegraphics[width=1in,height=1.25in,clip,keepaspectratio]{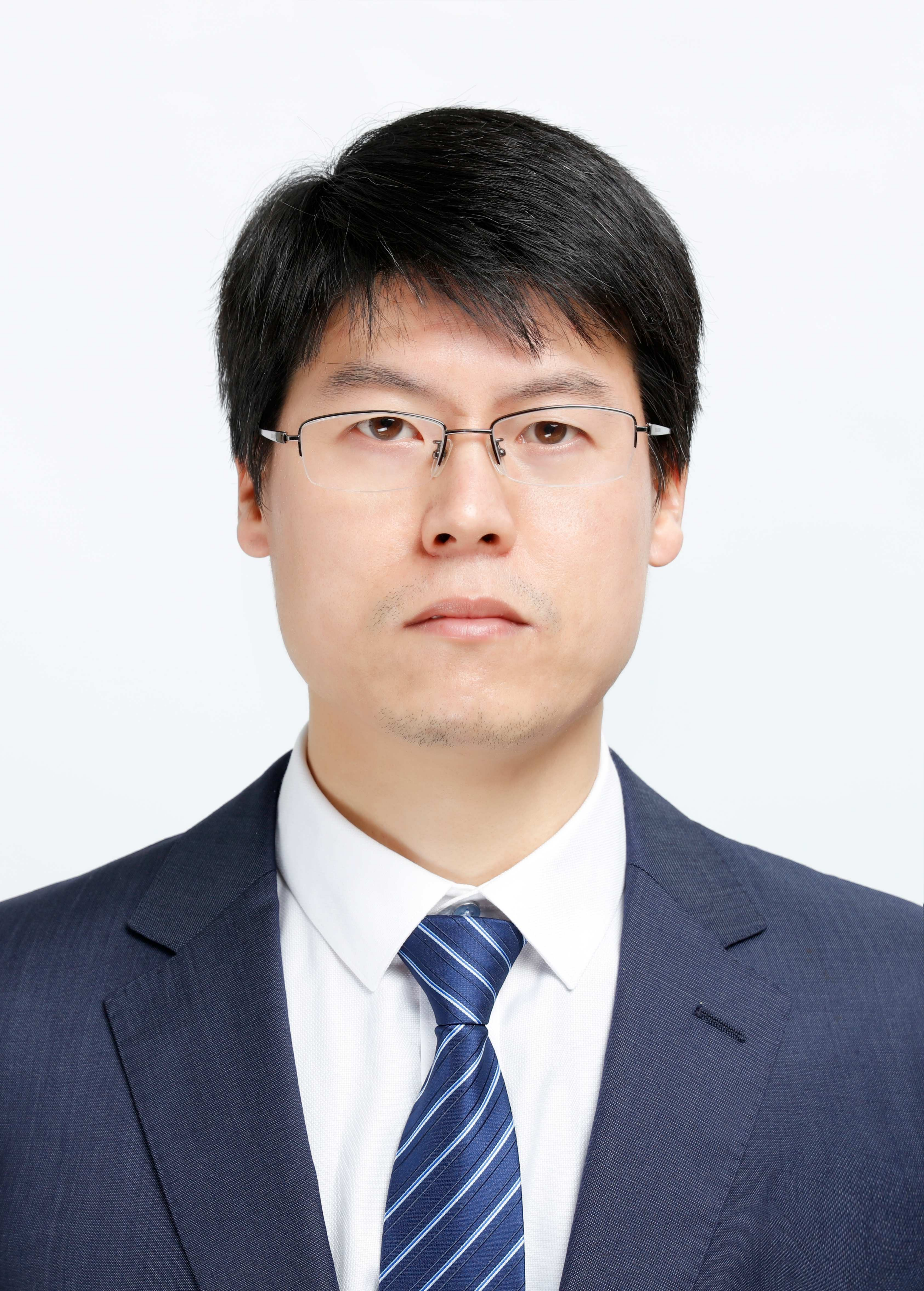}}]{Wen-An Zhang (Senior Member, IEEE)} received the B.Eng. degree in automation and the Ph.D. degree in control theory and control engineering from Zhejiang University of Technology, Hangzhou,China, in 2004 and 2010, respectively.\\ \indent From 2010 to 2011, he was a Senior Research Associate with the Department of Manufacturing Engineering and Engineering Management, City University of Hong Kong, Hong Kong. Since 2010,he has been with Zhejiang University of Technology, where he is currently a Professor with the Department of Automation. His current research interests include multi-sensor information fusion estimation and its applications.\\ \indent Dr. Zhang was awarded an Alexander von Humboldt Fellowship in 2011 and 2012. Since September 2016, he has been serving as a Subject Editor for Optimal Control Applications and Methods.
\end{IEEEbiography}
\begin{IEEEbiography}[{\includegraphics[width=1in,height=1.25in,clip,keepaspectratio]{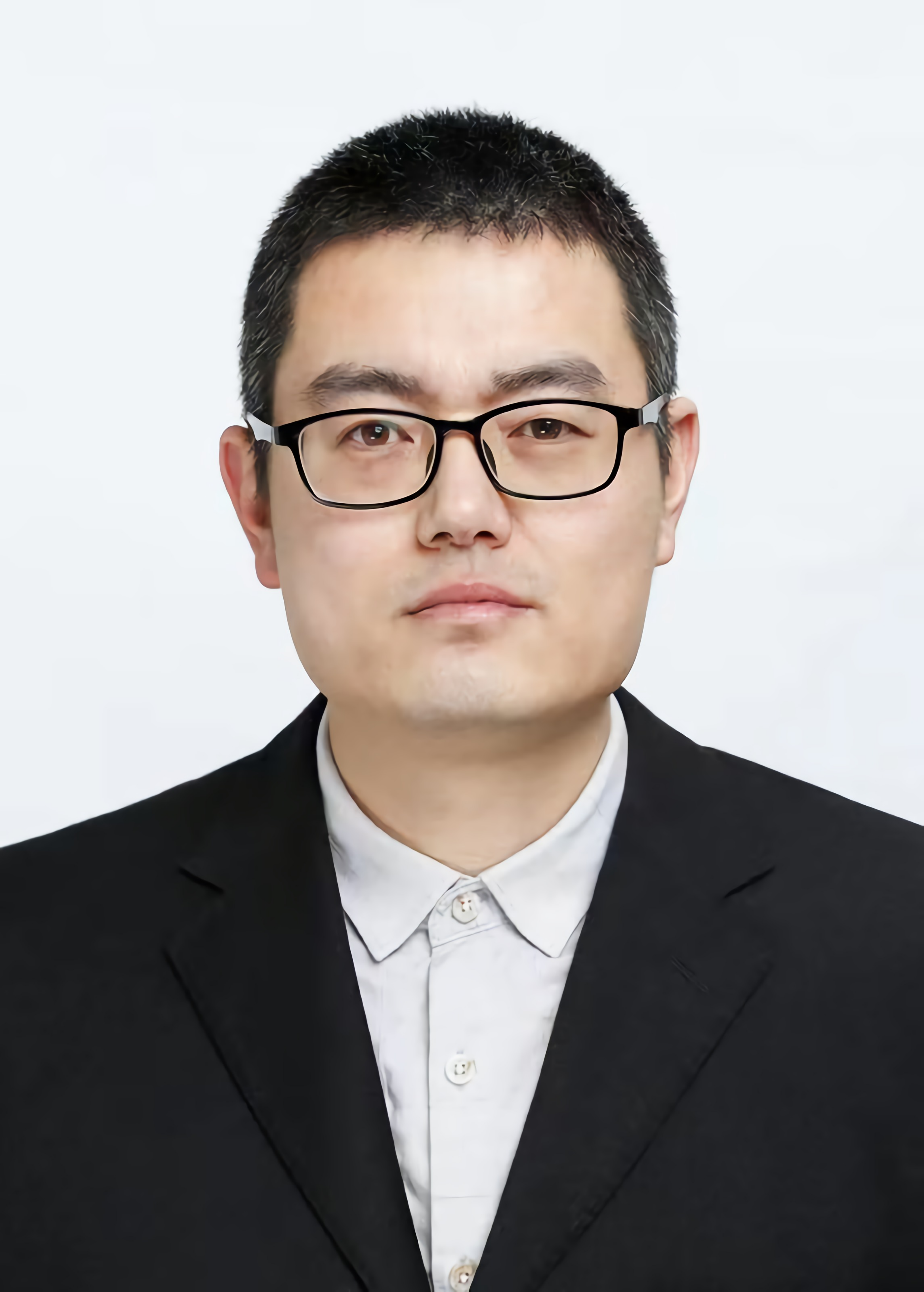}}]{Minglei Fu (Member, IEEE)} received the bachelor’s degree in communication engineering, the master’s degree in communication and information systems, and the Ph.D. degree in control theory and control engineering from Zhejiang University of Technology, Hangzhou, China, in 2004, 2007, and 2010,respectively.\\
\indent He is currently a Professor with the College of Information Engineering, Zhejiang University of Technology. In 2017, he was selected as the Discipline Leader of colleges and universities in Zhejiang province. He is the author of 80 articles and four monographs. He holds more than 20 invention patents. His research interests include computer vision and SLAM for mobile robots, environment perception, positioning, and navigation technology for autonomous aerial vehicles.
\end{IEEEbiography}

\end{document}